\documentclass{article}

\usepackage{microtype}
\usepackage{graphicx}
\usepackage{subcaption}
\usepackage{booktabs} % for professional tables

\usepackage{multirow}
\usepackage{multicol}
\usepackage{placeins}
\usepackage{rotating}
\usepackage{setspace}
\usepackage{enumitem}
\usepackage{textcomp}
\usepackage{threeparttable}
\usepackage{url}
\usepackage{verbatim}
\usepackage{wrapfig}
\usepackage{xcolor}
\usepackage{ulem}
\usepackage{colortbl}
\definecolor{mygray}{gray}{.9}

\usepackage[preprint]{icml2026}

\usepackage{amsmath}
\usepackage{amssymb}
\usepackage{mathtools}
\usepackage{amsthm}
\usepackage{comment}

\usepackage[capitalize,noabbrev]{cleveref}

\theoremstyle{plain}

\theoremstyle{definition}

\theoremstyle{remark}

\usepackage[textsize=tiny]{todonotes}

\begin{document}

\twocolumn[
  \icmltitle{STaRR: Spatial-Temporal Token-Dynamics-Aware Responsive Remasking for Diffusion Language Models}

  % It is OKAY to include author information, even for blind submissions: the
  % style file will automatically remove it for you unless you've provided
  % the [accepted] option to the icml2026 package.

  % List of affiliations: The first argument should be a (short) identifier you
  % will use later to specify author affiliations Academic affiliations
  % should list Department, University, City, Region, Country Industry
  % affiliations should list Company, City, Region, Country

  % You can specify symbols, otherwise they are numbered in order. Ideally, you
  % should not use this facility. Affiliations will be numbered in order of
  % appearance and this is the preferred way.
  \icmlsetsymbol{equal}{*}

  \begin{icmlauthorlist}
\icmlauthor{Xinhao Sun}{yyy}
    \icmlauthor{Huaijin Zhao}{yyy,comp}
    \icmlauthor{Zihao Zheng}{comp}
    \icmlauthor{Jiayu Chen}{sch}
    \icmlauthor{Maoliang Li}{yyy}
    \icmlauthor{Yun Liang}{sch,yyy,comp,corresponding}
    \icmlauthor{Xiang Chen}{comp,corresponding} % 标注通讯
    %\icmlauthor{}{sch}
    % \icmlauthor{Firstname8 Lastname8}{sch}
    % \icmlauthor{Firstname8 Lastname8}{yyy,comp}
    %\icmlauthor{}{sch}
    %\icmlauthor{}{sch}
  \end{icmlauthorlist}

  \icmlaffiliation{yyy}{Department of XXX, University of YYY, Location, Country}
  \icmlaffiliation{comp}{Company Name, Location, Country}
  \icmlaffiliation{sch}{School of ZZZ, Institute of WWW, Location, Country}

  \icmlcorrespondingauthor{Firstname1 Lastname1}{first1.last1@xxx.edu}
  \icmlcorrespondingauthor{Firstname2 Lastname2}{first2.last2@www.uk}

  % You may provide any keywords that you find helpful for describing your
  % paper; these are used to populate the "keywords" metadata in the PDF but
  % will not be shown in the document
  \icmlkeywords{Machine Learning, ICML}

  \vskip 0.3in
 ]

% this must go after the closing bracket ] following \twocolumn[ ...

% This command actually creates the footnote in the first column listing the
% affiliations and the copyright notice. The command takes one argument, which
% is text to display at the start of the footnote. The \icmlEqualContribution
% command is standard text for equal contribution. Remove it (just {}) if you
% do not need this facility.

% Use ONE of the following lines. DO NOT remove the command.
% If you have no special notice, KEEP empty braces:
\printAffiliationsAndNotice{}  % no special notice (required even if empty)
% Or, if applicable, use the standard equal contribution text:
% \printAffiliationsAndNotice{\icmlEqualContribution}

\begin{abstract}
\begin{comment}
% XH Original
The remasking strategy, which defers the decoding of low-priority tokens, is critical for efficiency. However, current remasking strategies are bottlenecked by using a single, global confidence threshold, failing to account for the unique temporal evolution and spatial correlation (i.e., the spatial-temporal dynamics) of individual tokens. This oversight introduces superfluous iterations and limits the potential for parallel generation.
We introduce an innovative remasking mechanism that overcomes this fixed-threshold limitation. Our method dynamically quantifies a token's convergence status and contextual relevance using two novel metrics: Temporal Variance and Spatial Deviance. Leveraging these dynamics, we implement token-wise, step-adaptive threshold adjustment.
Temporal Variance and Spatial Deviance. Leveraging these dynamics, we implement token-wise, step-adaptive threshold adjustment.
Temporal Variance and Spatial Deviance.
\end{comment}

Diffusion Language Models (DLMs) enable parallel decoding via iterative denoising, where remasking strategies play a critical role in balancing inference speed and output quality. 
Existing methods predominantly rely on static confidence thresholds, overlooking the spatial-temporal dynamics of token confidence, causing unnecessary remasking. 
We propose \textbf{\textit{Spatial-Temporal Token-Dynamics-Aware Responsive Remasking}} (STaRR), a training-free framework that dynamically adapts remasking decisions based on token confidence evolution. 
STaRR introduces two metrics, temporal variance and spatial deviance, to guide fine-grained, step-wise dynamic thresholding. 
We further introduce a step-wise dynamic thresholding strategy, further enhanced with responsiveness optimizations for scalability and robustness.
Experiments show that STaRR achieves an average speedup of $4.1\times$ and up to $8.9\times$ while maintaining comparable accuracy.

% Diffusion Language Models (DLMs) offer a promising alternative to autoregressive generation by enabling parallel decoding through an iterative denoising process. A key factor governing the efficiency and quality of DLM inference is the remasking strategy, which determines which low-confidence tokens are deferred to later diffusion steps. Existing approaches largely rely on static confidence thresholds, ignoring the spatial--temporal dynamics of token confidence and often leading to unnecessary remasking of semantically converged tokens. In this paper, we propose \textbf{\textit{Spatial--Temporal Token-Dynamics-Aware Responsive Remasking}} (STaRR), a training-free framework that accelerates DLM inference by dynamically adapting remasking decisions. STaRR systematically analyzes token confidence evolution during diffusion and quantifies it using two efficient metrics: temporal variance and spatial deviance. Based on these metrics, we introduce a fine-grained, step-wise dynamic thresholding strategy, further enhanced with responsiveness optimizations for scalability and robustness. Experiments on MBPP and GSM8K demonstrate that STaRR achieves an average speedup of $4.1\times$ and up to $8.9\times$ without sacrificing accuracy. Additionally, STaRR is compatible with existing acceleration techniques, enabling combined speedups exceeding $10\times$.

\end{abstract}

\section{\textbf{Introduction}}
\label{sec:intr}

Recently, Diffusion Language Models (DLMs)~\cite{wang2508diffusion,austin2021program} have emerged as a compelling alternative to autoregressive paradigms. 
DLMs show strong potential for unifying multimodal generation and benefit from parallel decoding, which enables high-throughput inference~\cite{wang2508diffusion,chen2021evaluating}. 
Instead of conducting next-token prediction, DLMs employ an iterative denoising process to transform unknown tokens gradually to answers.~\cite{yu2025discrete}.
Specifically, this iterative denoising process is achieved by a multi-step diffusion process.
%, which is compoised of Such an interative Therefore, the DLM decoding mechanism evolves from a fully masked input to a target sequence into a multi-step diffusion process.
And each diffusion step is divided into two phases: a prediction phase and a remasking phase.
During the prediction phase, the model generates candidate tokens along with confidence scores for all current masks.
During the remasking phase, a remasking strategy is applied to filter out low-confidence predictions, deferring their resolution to subsequent steps where richer contextual information enables more accurate inference.
The choice of tokens to remask is therefore critical, as it directly governs the trade-off between generation speed and output quality.

%Determining which tokens to remask based on the model’s output logits remains a challenging problem.
    A substantial body of prior work has focused on designing remasking strategies.
    LLaDA-8B~\cite{nie2025large} and Dream-7B~\cite{ye2025dream} adopt fixed-threshold strategies, decoding only tokens whose confidence exceeds a predefined cutoff.
    DUS~\cite{luxembourg2025plan} further shows that a token’s position can influence its decoding schedule.
    SlowFast Sampling~\cite{wei2025accelerating} extends this line of work by introducing constrained remasking, where a fast-decoding window is predicted at each step, and decoding is restricted to tokens within this window~\cite{tian2025finish}.

% However, most of these strategies are confined to utilizing static logit information. 
% As evidenced in Fig.~\ref{fig:1}, the token 'determine' is remasked simply because its confidence does not reach the threshold, but the model has captured the full meaning of this token. 
% Actually, this kind of issue is widely observed.
% Static Threshold can cause the remasking decisions to disconnect from the model's semantic understanding~\cite{chen2025dpad}. 
% To be specific, firstly, traditional strategies only use the instantaneous confidence of a token, thereby overlooking the model's dynamic process of evolving semantic understanding. Secondly, these strategies focus exclusively on local token-level information while neglecting the broader global context. 
% Due to the polysemy and abstraction of natural language, when using these remasking strategies, the whole decoding process can be delayed by a random uncritical token.
\begin{figure}[b]  % htbp是浮动位置参数
    \vspace{-7mm}
    \centering  % 居中显示
    \includegraphics[width=3.3in]{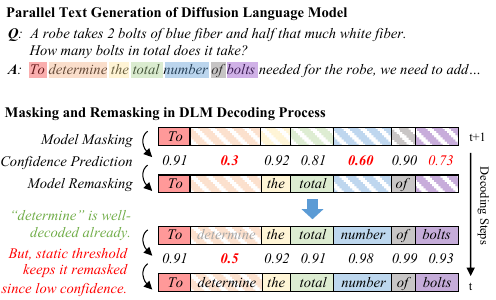}
    \vspace{-4mm}
    \caption{Limitations of Static Confidence in Diffusion Remasking}  % 图片标题
    \vspace{-2mm}
    \label{fig:1}  % 标签,用于交叉引用
\end{figure}

However, most existing strategies are restricted to static confidence thresholds for token remasking, even when spacial and temporal information is involved (e.g., DUS, SlowFast).
    As illustrated in Fig.~\ref{fig:1}, the example token pointing to the word of \textit{``determine''} is unnecessarily remasked due to its below-threshold confidence score, despite the model having already captured its semantic content.
    % Such behavior is commonly observed in practice.
    %Such issues can be caused  static threshodling existing approaches suffer from two key limitations.
    %First, they rely exclusively on the instantaneous confidence score of tokens, ignoring the dynamic evolution of semantic representations throughout the diffusion process.
    %Second, these strategies operate purely at the local token level and fail to account for global contextual information.
        %Thus, the static thresholding can lead to remasking decisions misaligned with the model’s underlying semantic understanding~\cite{chen2025dpad}, and delay the entire decoding process by uncritical tokens.
    %%Given the polysemy and abstract nature of natural language, such remasking strategies can cause the entire decoding process to be unnecessarily delayed by an uncritical token.
    Such issues arise fundamentally from the use of a static threshold.
    First, they rely exclusively on the instantaneous confidence scores of tokens, without considering the dynamic evolution of semantic representations throughout the diffusion process.
     Second, these strategies operate purely at the local token level and fail to incorporate global contextual information.
        Consequently, static thresholding may result in remasking decisions that are misaligned with the model’s underlying semantic understanding~\cite{chen2025dpad}, and unnecessarily delay the overall decoding process by repeatedly remasking low-uncertainty tokens.

Therefore, a better remasking strategy based on the actual dynamics of token confidence is needed:
    From a temporal perspective, token’s convergence trajectory exhibits distinct patterns that influence how its confidence evolves over time. It unveils the model's progressive semantic disambiguation at specific positions. Analyzing it provides insights into the underlying state of the token decoding process.
    From a spacial perspective, token confidence also shape inter-token distributions, which characterize the scope of the model's holistic linguistic comprehension. These distributions also facilitate the identification of critical areas.
Such dynamics provide critical insights into the model's token selection process and give proper suggestions for remasking.

\begin{figure}[t]  % htbp是浮动位置参数
    \centering  % 居中显示
    \includegraphics[width=3.3in]{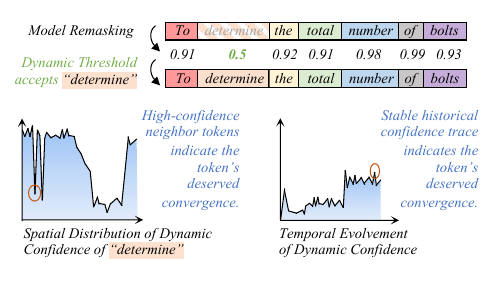}
    \vspace{-10mm}
    \caption{Token Remasking Strategy}  % 图片标题
    \vspace{-10mm}
    \label{fig:2}  % 标签，用于交叉引用
\end{figure}

\begin{comment}
% XH original
We analyzed these spacial-temporal dynamics of token confidence in detail and learned their underlying patterns. We also quantify these characteristics into two real-time computable metrics: temporal variance and spacial deviance.
Based on the analysis, a dynamic-threshold algorithm is developed. 
In each step, we calculate a specific dynamic threshold for each token based on its temporal variance and spacial deviance. It will be used to guide the remasking process.
We then established a framework of \textbf{\textit{spacial-Temporal Token-Dynamics-Aware Responsive Remasking}} (STaRR),  a training-free method for accelerating inference.
By incorporating Responsiveness Optimization (RO), STaRR naturally adapts to varying context lengths, prompt difficulties, and generation styles. Fig.~\ref{fig:2} demonstrates how, by incorporating spacial-temporal dynamics, our approach successfully distinguishes tokens that are 'converged yet low-confidence' from truly uncertain ones. We also conduct extensive experiments and compare our method with the state-of-the-art remasking strategies.
\end{comment}

In this work, we introduce \textbf{\textit{spacial--Temporal Token-Dynamics-Aware Responsive Remasking}} (STaRR), a training-free framework for accelerating DLM inference.
STaRR is built upon a detailed analysis of the spacial-temporal dynamics of token confidence.
    It quantifies these dynamics into two real-time, computationally efficient metrics: temporal variance and spacial deviance.
    Leveraging these metrics, a dynamic threshold remasking strategy is proposed at a fine granularity of each decoding step.
    Furthermore, STaRR is enhanced by a set of responsiveness optimization methods for its scalability and robustness.
    %adaptively adjusts to varying context lengths, prompt difficulties, and generation styles.

As illustrated in Fig.~\ref{fig:2}, STaRR can effectively prevent tokens that are semantically converged but below a static confidence score from unnecessarily remasking as occurred in Fig.~\ref{fig:1}.
Experimental results on popular benchmarks, including MBPP and GSM8K, show that our method achieves an average speedup of $4.1\times$ and a maximum speedup of up to $8.9\times$, while maintaining comparable answer accuracy.
Moreover, our approach is compatible with many existing acceleration techniques, such as dKV-Cache and in-place prompting, enabling combined speedups exceeding $10\times$.
    %Extensive experiments demonstrate that STaRR consistently outperforms state-of-the-art remasking strategies across multiple benchmarks.

\begin{comment}
Our contribution can be summarized as follows:
\vspace{-2mm}
\begin{itemize}[leftmargin=*]
    \vspace{-2mm}
    \item We introduce novel spacial-temporal metrics to characterize token dynamics during diffusion decoding.
    \vspace{-2mm}
    \item We escalate the remasking strategy from fixed-threshold heuristics to a dynamic manner, allowing more flexible token selection in diffusion language models.
    \vspace{-2mm}
    \item We propose corresponding a training-free acceleration framework that significantly improves inference efficiency of diffusion language models without sacrificing generation quality.
\end{itemize}
\vspace{-2mm}
Experimental results on popular benchmarks, including MBPP and GSM8K, show that our method achieves an average speedup of $4.1\times$ and a maximum speedup of up to $8.9\times$, while maintaining comparable answer accuracy.
Moreover, our approach is compatible with many existing acceleration techniques, such as dKV-Cache and in-place prompting, enabling combined speedups exceeding $10\times$.
\end{comment}
\section{\textbf{Prelimilary}}
\label{sec:prel}

% ==============================================================================
% subsection
% ==============================================================================
\subsection{\textbf{Diffusion Language Models}}
Diffusion language models (DLMs) extend diffusion-based generation to discrete text by replacing left-to-right autoregressive decoding with iterative refinement over the entire sequence.
    Given a fixed-length sequence $x$, in which all tokens are masked except for the prompt, the model first predicts tokens for all positions in parallel using a trained mask predictor.
    It then applies a remasking mechanism that preserves high-confidence tokens while reverting the other tokens to masks.
    This process is repeated until all tokens are decoded or the number of iterations exceeds a predefined maximum~\cite{chen2025dpad,kim2025rainbow}.
    The update procedure at step $t$ can be described as follows:
\begin{equation}
\label{eq:vector_mapping}
\begin{aligned}
    \begin{bmatrix}
    \overline{x}_1 \\
    \overline{x}_2 \\
    \vdots \\
    \overline{x}_n
    \end{bmatrix}
        \xrightarrow{model}
    \begin{bmatrix}
    (x_1, c_1) \\
    (x_2, c_2) \\
    \vdots \\
    (x_n, c_n)
    \end{bmatrix}
        \xrightarrow[confidence\ threshold\ \tau]{Remasking\ Strategy}
    \begin{bmatrix}
    \overline{x}_1 \\
    x_2 \\
    \vdots \\
    \overline{x}_n
    \end{bmatrix}
\end{aligned},
\end{equation}
where $x_i$ denotes the decoded token at position $i$, and $\overline{x}_i$ denotes a masked token.
The remasking strategy determines, at each step, which tokens to remask based on the token-wise confidence scores $c_i$ produced by the model.

% ==============================================================================
% subsection
% ==============================================================================
\subsection{\textbf{Remasking Strategies in DLM}}
The remasking process plays a critical role in DLMs, as it alleviates the burden on the model of determining all token identities within a single iteration~\cite{arriola2025block}.
    At each step, a remasking strategy must select an appropriate subset of tokens to decode based on the model’s output logits, which in turn requires reliable and dynamically updated confidence estimates.
    However, existing remasking strategies generally fail to account for both the temporal and spacial dynamics of token confidence.

Some approaches rely on static, step-independent confidence measures.
    A representative example is the fixed-threshold strategy, which predefines a constant confidence threshold $\tau$.
    At each decoding step, tokens whose confidence exceeds $\tau$ are retained, while those with confidence below the threshold are remasked.

Other approaches adopt heuristic remasking strategies that incorporate temporal information but neglect spacial dynamics.
    This strategy adjusts the confidence threshold as a function of the decoding step, thereby integrating temporal progression into the remasking decision.
    Methods such as TSE~\cite{wang2025time}, Rainbow~\cite{kim2025rainbow}, and Saber~\cite{dong2025saber} fall into this category.

In contrast, some approaches exploit spacial structure while ignoring temporal dynamics.
    DUS (Dilated Unmasking Strategy)~\cite{luxembourg2025plan} partitions sequence positions into non-adjacent dilated groups and unmasks them in parallel to minimize an upper bound on the joint entropy gain at each denoising step.
    Similarly, Block Diffusion~\cite{arriola2025block,lu2025adablock} and SlowFast Sampling~\cite{wei2025accelerating} constrain the decoding region at each step and remask the remaining positions according to predefined or heuristic rules.

\section{\textbf{Temporal-spatial Token Dynamics}}
\label{sec:3}
\begin{figure}[!t]  % htbp是浮动位置参数
    \centering  % 居中显示
    \includegraphics[width=3.3in]{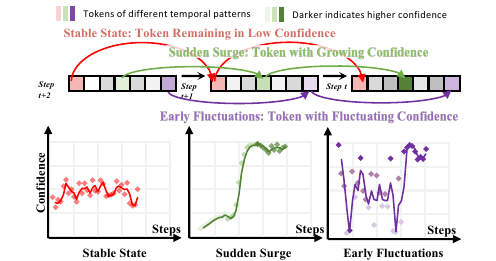}
    \vspace{-2mm}
    \caption{Token Patterns Categorized by Temporal Variance}  % 图片标题
    \vspace{-4mm}
    \label{fig:3}  % 标签，用于交叉引用
\end{figure}

To develop a principled remasking strategy that jointly exploits token confidence and spatial-temporal dynamics, we first conduct an empirical analysis of confidence trajectories in DLM decoding.
Specifically, we examine:
(i) how the confidence of each token evolves across diffusion steps (\textit{temporal dynamics}), and
(ii) how token confidence is distributed throughout the sequence at intermediate decoding steps (\textit{spatial dynamics}).

% ==============================================================================
% subsection
% ==============================================================================
\subsection{\textbf{Temporal Analysis for Specific Token Confidence}}

\textbf{\textit{Pattern Analysis of Temporal Dynamics}}: 
% ========== Original XH's text ==========
% Token-level temporal variance exhibits substantial heterogeneity. Examining it across datasets yields an empirical distribution.
% Specifically, we discover three main temporal patterns during the DLM decoding process, as shown in Fig.~\ref{fig:3}:
%The temporal variance exhibits substantial heterogeneity.
Through analysis of confidence trajectories during diffusion decoding, we identify three dominant temporal patterns to summarize the temporal dynamics as illustrated in Fig.~\ref{fig:3}:

\vspace{-2mm}
\hspace{4mm}\textit{Stable State}: Confidence remains approximately constant around a fixed value over a long period.
This may occur before decoding, where confidence is typically low, or after decoding, where it stabilizes at a high level.

\vspace{-2mm}
\hspace{4mm}\textit{Sudden Surge}: 
Confidence surges sharply within a small number of steps before transitioning into another plateau.
This phase does not persist for many iterations.

\vspace{-2mm}
\hspace{4mm}\textit{Early Fluctuations}: Confidence values fluctuate significantly in recent steps and fail to stabilize. This often occurs in the early steps of decoding.

% ========== Original XH's text ==========
% Most tokens follow a trajectory that progresses from early oscillations to a low-level plateau, followed by a sudden surge, and finally reaches a high-level stable state. 
%     However, the timing of these patterns is highly asynchronous across different tokens, and some may even skip certain stages entirely. 
%     By analyzing the specific temporal pattern of a token’s confidence, we can discern whether a token is currently understood by the model or if it possesses the potential to be understood in future steps.
Most tokens follow a trajectory that transitions from early fluctuations to a low-confidence plateau, followed by a sudden surge, and finally converges to a high-confidence stable state.
    However, the occurrence of these phases is highly asynchronous across tokens, and some tokens may bypass certain stages.
    By characterizing the temporal confidence pattern of each token, it is possible to predict whether the token has already been well understood by the model.

\textbf{\textit{Quantitative Formulation of Temporal Dynamics}}:

Having identified distinct patterns in the temporal dynamics of confidence, we now introduce a quantitative metric to enable a more rigorous analysis.
Specifically, we define \textbf{\textit{temporal variance}}, which captures the deviation of a token’s current confidence from its previous steps.
This value is determined by calculating the residual between the current confidence value and the weighted average of confidence scores from previous steps.
Formally, for each position $i$, let $c_t^i$ denote the confidence score at decoding step $t$.
    To characterize the convergence process of individual tokens, we define the \textbf{\textit{temporal variance}} as Eq.~\ref{eq:2}:
\begin{equation}
    var_{t}^{i}(W_{t}) = 
        c_{t}^{i}-\frac{1}{W_{t}}\sum_{j=t-W_{t}}^{t-1}{c_{j}^{i}}
\label{eq:2}
\end{equation}
To improve computational efficiency and to exclude distant, less-correlated states, we introduce a temporal window that constrains the computation of temporal variance.
As a result, only confidence information within this localized temporal context influences the adaptive decision-making at the current step.
Here, $W_t$ denotes the window size.
While a larger $W_t$ provides a richer temporal signal, under a finite number of denoising steps $T_{\max}$, it may introduce lag and yield stale estimates.
Notably, even after a token has been decoded, its confidence may continue to fluctuate within a narrow range, so it still has its unique variance value.

\textbf{\textit{Optimization Insights}}:
A high temporal variance typically indicates that a token is actively being resolved by the model.
    In particular, such a token may be experiencing Early Fluctuations or a Sudden Surge.
    At this point, committing to decoding is premature, even if the token has already surpassed a predefined confidence threshold.
Conversely, a low temporal variance suggests that the model has incorporated little new information about the token in recent steps.
    Under this condition, decoding the token is more appropriate.
    However, to avoid decoding tokens that are on the verge of a Sudden Surge, it is necessary to further examine their spatial dynamics as complementary verification.

These observations imply that employing a fixed decoding threshold throughout the generation process is inherently suboptimal.
    Instead, decoding thresholds should be adaptively determined at each timestep, in accordance with their generation characteristics.

% ==============================================================================
% subsection
% ==============================================================================

\begin{figure}[!t]  % htbp是浮动位置参数
    \centering  % 居中显示
    \includegraphics[width=3.3in]{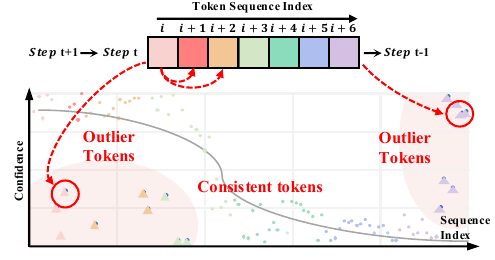}
    \vspace{-4mm}
    \caption{Token Patterns Categorized by Spatial Deviance}  % 图片标题
    \vspace{-4mm}
    \label{fig:4}  % 标签，用于交叉引用
\end{figure}

\subsection{\textbf{Spatial Analysis for Specific Decoding Step}}

\textbf{\textit{Pattern Analysis of Spatial Dynamics}}: spatial deviance also exhibits substantial variability across tokens.
Fig.~\ref{fig:4} illustrates the correlation of token confidence at a given step, from which two dominant spatial patterns emerge.

\vspace{-2mm}
\hspace{4mm}\textit{Consistent}: The confidence level shows a minimal gap compared to that of its neighboring tokens.

\vspace{-2mm}
\hspace{4mm}\textit{Outlier}: The token’s confidence differs markedly from its spatial context. It may serve as a ``high-confidence anchor'' in a low-confidence region, or persist as a ``low-confidence outlier'' despite high-confidence neighbors.

At most decoding steps, token confidence is consistent with its neighbors.
However, when the decoding progress of a particular token substantially leads or lags behind that of its neighbors, its confidence manifests as a spatial outlier.

\textbf{\textit{Quantitative Formulation of Spatial Dynamics}}:
We have revealed the distinct patterns of spatial dynamics.
To formalize the analysis of these patterns, we introduce a novel metric termed \textbf{\textit{spatial deviance}}.
It represents the degree of consensus between a token's confidence and that of its spatial neighbors.
It is calculated as the residual between a token’s own confidence and the weighted average of its neighbors' confidence scores.
\begin{equation}
dev_{t}^{i}(\mathcal{N}_{i, \text{in}}) =
    c_{t}^{i} - \sum_{j \in \mathcal{N}_{i, \text{in}}} w_{x, j}^{\text{norm}} \cdot c_{t}^{j}
\label{eq:3}
\end{equation}
Similarly, motivated by the principle of spatial locality, we define a spatial observation window for each token.
    This constraint ensures that only the confidence scores of tokens within a localized neighborhood are used to compute the \textbf{\textit{spatial deviance}} at a given position.
    In Eq.~\ref{eq:3}, $\mathcal{N}_{i,\text{in}}$ denotes the spatial window we look into.
    A larger window incorporates richer contextual information, but its size must remain limited given the finite sequence length.
    The term $w_{x,j}^{\text{norm}}$ represents the weight assigned to each surrounding token, typically decreasing with spatial distance from the target token so that closer neighbors exert greater influence.
    % The sum over time of $|dev_{t}^{i}(\mathcal{N}_{i, \text{in}})|$ (whole deviance of a token) reveals the isolation of token $i$ throughout the decoding process.
    Aggregating the spatial deviance $|dev_t^i(\mathcal{N}_{i,\text{in}})|$ over time, referred to as the whole deviance of a token, reveals the extent to which token $i$ remains isolated from its surrounding context throughout the decoding process.

\textbf{\textit{Optimization Insights}}:
Model understanding is often regional; thus, a high spatial deviance indicates that a token’s remasking decision is inconsistent with its local context.
Such inconsistency catalyzes the token to update its state, potentially transitioning from \textit{remask} to \textit{decode}, or vice versa.
In contrast, a low deviance suggests that the token’s current status should be preserved.

Given the heterogeneous spatial dynamics exhibited by different tokens, adopting a uniform decoding threshold is fundamentally inappropriate.
Instead, decoding thresholds should be adapted on a per-token basis to account for these localized disparities.

\begin{figure*}[!t]  % htbp是浮动位置参数
    \centering  % 居中显示
    \includegraphics[width=7in]{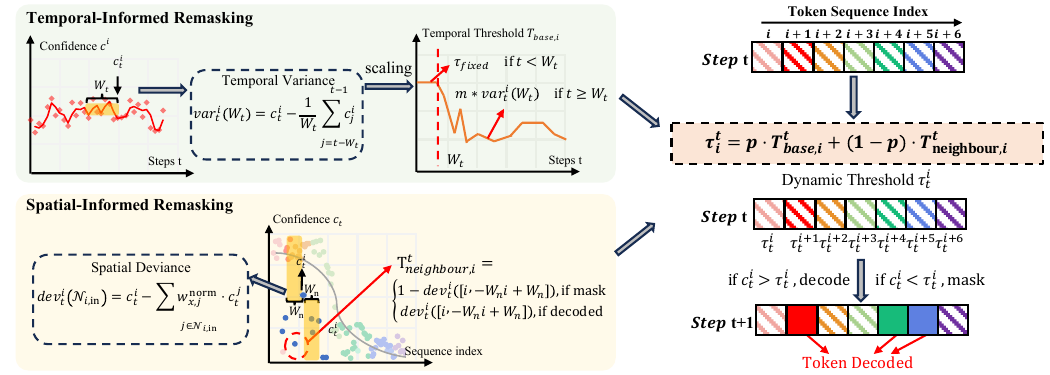}
    \vspace{-5mm}
    \caption{Overview of Dynamic Token Remasking for Diffusion Language }  % 图片标题
    \vspace{-2mm}
    \label{fig:5}  % 标签，用于交叉引用
\end{figure*}

\subsection{STaRR Design Motivation}
To effectively leverage the properties of token dynamics, we propose \textit{\textbf{STaRR: spatial-Temporal Token-Dynamics-Aware Responsive Remasking}} for
Diffusion Language Models.
The entire remasking framework consists of two components: Dynamic Threshold Remasking and Responsiveness Optimization. They work together to make each token decoded correctly and in time.

% As illustrated in Fig.~\ref{fig:6}, 
The remasking process begins by computing dynamic thresholds for all tokens using their temporal variance and spatial deviance. Each token is then evaluated for decoding by comparing its current confidence with the corresponding dynamic threshold. Detailed information about Dynamic Threshold Remasking will be introduced in Section.~\ref{sec:4}.
Subsequently, the sequence is passed to the Responsiveness Optimization mechanism, which further refines the remasking through a token-labeling strategy. Section.~\ref{sec:5} provides more information about Responsiveness Optimization.

\section{\textbf{Dynamic Threshold Remasking}}
\label{sec:4}

%\subsection{Confidence Threshold Adjustment Overview}
Dynamic Threshold Remasking leverages the spatial-temporal dynamics of tokens to establish individualized thresholds, thereby integrating these dynamics into the remasking process.
Fig.~\ref{fig:5} demonstrates how we assign dynamic confidence thresholds for each token at every step, and how we make remasking decisions by comparing specific thresholds with token confidence.

The threshold for token $i$ in step $t$ is determined by two terms:
$T_{base, i}^{t}$ and $T_{neighbour, i}^{t}$, which are respectively related to the temporal and spatial deviance:
\begin{equation}
    \tau_{t}^{i} = p *T_{base,i}^{t} + (1-p)*T_{neighbour,i}^{t}
\label{eq:4}
\end{equation}
% The spatial-temporal modulation factor $p$ determines the relative importance between the temporal and spatial factors. The two metrics are integrated using a straightforward linear formulation.
% In the early stages, when there is less historical information, greater emphasis should be placed on spatial information ($\text{T}_{neighbour, i}^{t}$). In the intermediate stages, the token's historical information ($\text{T}_{base, i}^{t}$) becomes more crucial, as it indicates the token's current stage. Finally, in the late stages, most tokens have accumulated sufficient information, and spatial characteristics once again become important. 
    % Consequently, the value of $p$ should exhibit a three-stage, step-like change: low, then high, then low again, as the decoding steps progress. In practice, we use the proportion of decoded tokens to determine which stage of decoding it is and choose a proper parameter $p$.
The spatial-temporal modulation factor $p$ determines the relative importance between temporal and spatial factors, which are combined through a simple linear formulation. 
In the early steps, when limited historical information is available, greater emphasis should be placed on spatial information ($\text{T}_{\text{neighbour}, i}^{t}$). 
During the intermediate steps, the token’s historical information ($\text{T}_{\text{base}, i}^{t}$) becomes more crucial, as it reflects the token's current convergence state. 
In the later steps, once most tokens have accumulated sufficient historical evidence, spatial characteristics again play a more prominent role.
The calculation of each token's threshold is parallel with model inference, so our method introduces negligible overhead.

% ==============================================================================
% subsection
% ==============================================================================
\subsection{\textbf{Temporal-Informed Remasking}}
Based on the observations in Section~\ref{sec:3}, tokens with lower temporal variance are more likely to have reached convergence. 
Accordingly, we assign these tokens a lower confidence threshold to enable earlier decoding, which can prevent token stalling.
In contrast, tokens with higher temporal variance require a more conservative treatment and are therefore associated with a higher threshold. 
For tokens lacking sufficient historical observations, like tokens in early steps, we temporarily adopt a fixed threshold as a fallback. 
Formally, the \textbf{\textit{temporal threshold}} is defined as follows:
\begin{equation}
    \text{T}_{base, i}^{t} =
        \begin{cases}
            \tau_{fixed} & \text{if } t<W_{t} \\
            m*var_{t}^{i}(W_t) & \text{if } t\geq W_{t} \\
        \end{cases}
\label{eq:5}
\end{equation}
where $\tau_{fixed}$, $m$, and $W_{t}$ are all tunable parameters for trade-offs between decoding speed and generation quality.

\textbf{\textit{Base Threshold for Warm-up Steps}}: $\tau_{\text{fixed}}$ controls the number of tokens decoded during the warm-up phase and does not affect subsequent decoding stages. To ensure conservative behavior, we set $\tau_{\text{fixed}}$ to $0.9$ by default, following the standard threshold used in Fast-dLLM.

\textbf{\textit{Temporal Threshold Window Size}}: $W_t$ dictates the maximum number of preceding steps considered when analyzing temporal variance. It should be chosen based on the expected number of decoding steps. Intuitively, a longer decoding horizon calls for a larger window size. In our main experiments, we set $W_t = 5$. Additional ablation studies on this parameter are provided in the appendix.

\textbf{\textit{Threshold Scaling}}: $m$ is a sensitive hyperparameter that governs the aggressiveness of decoding. 
    When relying solely on the temporal-variance-based threshold, a token is decoded only if $c_t^i > T_{\text{base}}^i$, which implies $c_t^i < \frac{m}{m-1} \cdot \bar{c}_{\text{hist}}^i$ where $\bar{c}_{\text{hist}}^i$ denotes the weighted historical confidence. 
    A smaller value of $m$ results in more conservative decoding behavior, whereas a larger $m$ accelerates the decoding process.
    Empirically, we find that setting $m = 3$ achieves a favorable trade-off between generation speed and output quality.

Overall, $T_{\text{base}, i}^t$ regularizes confidence convergence by encouraging early decoding of stable tokens while remasking those exhibiting high uncertainty.
This mechanism allows the model to focus computation on uncertain positions, thereby improving the efficiency of the overall generation.

% ==============================================================================
% subsection
% ==============================================================================
\subsection{\textbf{Spatial-Informed Remasking}}
% Beyond temporal behavior, a token's stability is also influenced by its relationship with neighboring tokens.
% To formalize this intuition, we define a new threshold $\text{T}_{neighbour, i}^{t}$ using a token's spatial deviance.
% To calculate it, we delineate a Neighbor Window of size $W_n$, and the range $(x-W_n, x+W_n)$ serves as the spatial window $\mathcal{N}_{i,\text{in}}$. 
% An exponential weighting scheme is employed. For example, for a window size of $W_n=3$, the weights $w_d$ assigned to tokens at distances $d=1, 2, 3$ are $w_1=1/4$, $w_2=1/8$, and $w_3=1/8$, respectively. 
% The \textit{spatial threshold} calculated from these parameters is designated as $\text{T}_{neighbour,i}^{t}$, as implied in Eq.~\ref{eq:6}.

Beyond temporal dynamics, a token’s stability is also influenced by its interactions with neighboring tokens. 
To formalize this intuition, we introduce a spatially informed threshold, denoted as $\text{T}_{\text{neighbour}, i}^{t}$, as defined in Eq.~\ref{eq:6}, based on the token’s spatial deviance. 
Specifically, we define a neighbor window of size $W_n$, where the interval $(x - W_n, x + W_n)$ constitutes the spatial neighborhood $\mathcal{N}_{i,\text{in}}$. 
An exponential weighting scheme is applied within this window to emphasize closer neighbors. For instance, when $W_n = 3$, the weights $w_d$ assigned to tokens at distances $d = 1, 2, 3$ are $w_1 = 1/4$, $w_2 = 1/8$, and $w_3 = 1/8$, respectively.

Spatial threshold indicates whether a token should alter its remasking condition. To be specific, we define the threshold based on two distinct scenarios. For a token in the ``masked'' state, a high spatial deviance suggests it should transition to ``decoded'' more easily; thus, we assign it a lower threshold. Conversely, for a token already in the 'decoded' state, the opposite logic applies. While traditional DLM decoding typically treats decoded tokens as final, our method allows these tokens to turn back to mask at any step.

\begin{equation}
\text{T}_{neighbour, i}^{t} = 
\begin{cases}
\tau_{start,i} , \text{if $i-W_{n}<$ Prompt Len.}  \\
\tau_{end,i} , \text{if $i+W_{n}>$ Sequence Len.}  \\
1-dev_{t}^{i}([i-W_{n},i+W_{n}]) , \text{if mask.} \\
dev_{t}^{i}([i-W_{n},i+W_{n}]) , \text{if decoded.} \\
\end{cases}
\label{eq:6}
\end{equation}

\textbf{\textit{Base Threshold of Sequence}}: 
In our method, we use the padding strategy for the spatial edge conditions. $\tau_{start, i}$ and $\tau_{end, i}$ are calculated using the same equation as other tokens in the middle, with a fixed padding confidence on the left or right. How we pad the prompt, and the end of text, means how much we can trust the ``virtual'' decoded tokens on the left and right side of the sequence. Usually, the prompt, which is on the left side, is more reliable, so we pad it with $1$, and those beyond the right end are unknown, so we pad it with $0$. A higher right-padding may encourage the "right-to-left" decoding in diffusion language models.

\textbf{\textit{Spatial Threshold Window Size}}: $W_{n}$ determines the extent of the neighborhood context considered around each target token. It should be selected according to the length of the sequence. Longer sequences need a larger window for a broader view of global information. 
% ${T}_{neighbour, i}^{t}$ as part of the dynamic threshold helps make token correlation more consistent. Combined with ${T}_{base, i}^{t}$, it can help detect and judge the slow tokens and decode them in time for the overall efficiency. At the same time, it will raise the dynamic confidence threshold for erratic tokens to prevent them from being decoded prematurely.

As a component of the dynamic thresholding mechanism, ${T}_{\text{neighbour}, i}^{t}$ promotes more consistent modeling of token correlations. When combined with the temporal threshold ${T}_{\text{base}, i}^{t}$, it facilitates the identification of slowly converging tokens and enables their timely decoding. Simultaneously, this mechanism raises the confidence threshold for erratic tokens, preventing them from being decoded prematurely.

\begin{table*}[!t]
\centering
\scriptsize
\setlength{\tabcolsep}{4.5mm} % 根据页面宽度可微调此数值
\caption{Overall performance comparison across four benchmarks.}
\vskip -0.05 in
\begin{tabular}{c|c|cccc|cccc}
\toprule
\toprule
\multirow{2}{*}{\textbf{Dataset}} & \multirow{2}{*}{\textbf{Metric}} & \multicolumn{4}{c|}{\textbf{LLaDA-8B}} & \multicolumn{4}{c}{\textbf{Dream-7B}} \\
\cmidrule{3-10}
~ & ~ & \textbf{Conf.} & \textbf{Fast.} & \textbf{DUS} & \textbf{STaRR} & \textbf{Conf.} & \textbf{Fast.} & \textbf{DUS} & \textbf{STaRR} \\
\midrule
\multirow{3}{*}{Gsm8k} & Accuracy $\uparrow$ & 78.2 & 75.8 & 72.1 & \textbf{73.6} & 75.0 & 74.2 & 65.2 & \textbf{75.0} \\
~ & TPS $\uparrow$ & 5.3 & 14.3 & 14.3 & \textbf{16.32} & 8.9 & 14.1 & 24.2 & \textbf{0.44} \\
~ & Speedup $\uparrow$ & 1.00$\times$ & 2.70$\times$ & 2.70$\times$ & \textbf{3.07$\times$} & 1.00$\times$ & 1.58$\times$ & 2.72$\times$ & \textbf{3.41$\times$} \\
\midrule
\multirow{3}{*}{MBPP} & Accuracy $\uparrow$ & 29.4 & 28.4 & 29.2 & 28.2 & 56.6 & 53.8 & 46.4 & \textbf{57.2} \\
~ & TPS $\uparrow$ & 5.4 & 22.4 & 14.6 & \textbf{48.1} & 10.2 & 30.2 & 27.3 & \textbf{0.32} \\
~ & Speedup $\uparrow$ & 1.00$\times$ & 4.15$\times$ & 2.70$\times$ & \textbf{8.90$\times$} & 1.00$\times$ & 2.96$\times$ & 2.68$\times$ & \textbf{2.91$\times$} \\
\midrule
\multirow{3}{*}{MATH500} & Accuracy $\uparrow$ & 33.4 & 33.4 & 21.4 & \textbf{35.1} & 38.4 & 37.9 & 27.0 & \textbf{37.9} \\
~ & TPS $\uparrow$ & 8.8 & 23.2 & 23.6 & \textbf{32.91} & 11.2 & 27.1 & 30.3 & \textbf{40.9} \\
~ & Speedup $\uparrow$ & 1.00$\times$ & 2.63$\times$ & 2.68$\times$ & \textbf{3.74$\times$} & 1.00$\times$ & 2.42$\times$ & 2.71$\times$ & \textbf{3.65$\times$} \\
\midrule
\multirow{3}{*}{HumanEval} & Accuracy $\uparrow$ & 40.8 & 37.1 & 13.2 & \textbf{30.2} & 28.2 & 49.2 & 12.2 & \textbf{29.7} \\
~ & TPS $\uparrow$ & 16.0 & 32.9 & 15.2 & \textbf{34.8} & 11.5 & 15.6 & 7.1 & \textbf{21.9} \\
~ & Speedup $\uparrow$ & 1.00$\times$ & 2.05$\times$ & 0.97$\times$ & \textbf{2.18$\times$} & 1.00$\times$ & 1.35$\times$ & 0.61$\times$ & \textbf{1.90$\times$} \\
\bottomrule
\bottomrule
\end{tabular}
\label{tab:1}
\end{table*}
\section{\textbf{Responsiveness Optimization}}
\label{sec:5}
\begin{figure}[!t]  % htbp是浮动位置参数
    \centering  % 居中显示
    \includegraphics[width=3.3in]{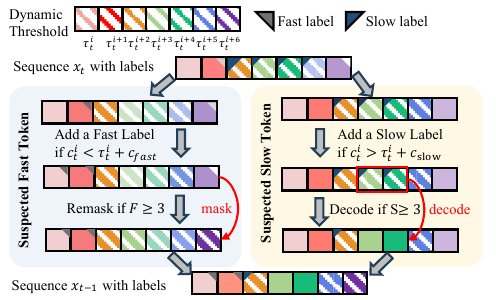}
    \vspace{-3mm}
    \caption{Token Patterns Categorized by Temporal Variance}  % 图片标题
    \vspace{-8mm}
    \label{fig:6}  % 标签，用于交叉引用
\end{figure}
Dynamic Threshold Remasking is a good way to combine the token's temporal variance and spatial deviation for a more accurate remasking.
However, extreme scenarios may still occur because in such a scene, a minor difference in confidence can lead to a drastic change in the decoding result, which may lead to a collapse of the model's answer. This is mainly because merely adjusting the threshold only indirectly influences the decoding process. 

For robustness,
Since a standalone dynamic threshold remasking mechanism cannot guarantee that every remasking decision is correct at each step, we propose a cross-step verification mechanism. This approach labels tokens that are potentially erroneous, allowing the remasking strategy to rectify previous mistakes in subsequent steps.

For scalability, variations in prompt length and reasoning complexity can impact the stability of the dynamic threshold. Incorporating an inter-step regulatory metric helps the model adapt to a wide variety of problem scenarios.

Responsiveness Optimization gives a confidence cushion
for all the tokens. And it allows temporal-spatial information to be transferred across steps using token labels. The metric primarily comprises two strategies: Suspected $\text{Fast}$ Token Mechanism and Suspected $\text{Slow}$ Token Mechanism. As illustrated in Fig.~\ref{fig:6}, we maintain two types of labels for state management. At the end of each step, if a token's confidence is in close proximity to its corresponding threshold, a ``fast'' or ``slow'' label is assigned based on its current decoding status. Conversely, the respective label is removed. Once the accumulation of either label reaches a threshold of three, the token undergoes a transition in its decoding state.

% This metric aims to identify tokens with extreme dynamics that might be overlooked by dynamic thresholding. By monitoring their decoding trajectories, it enables the model can directly intervene and determine their decoding status at the optimal moment. It mainly labels two kinds of tokens.

\subsection{Suspected $\text{Fast}$ Token Mechanism} 

In the initial $t_{\text{start}}$ steps, if a token is decoded with a confidence exceeding its threshold by no more than $c_{\text{fast}}$, it is marked with a \textit{``Suspected Fast''} label. 
    The token corresponding to the highest-confidence content is then closely monitored. 
    If the content changes, the token is immediately remasked to avoid premature decoding. 
    If the content remains unchanged throughout the $t_{\text{start}}$ steps, the label is removed. 
    This mechanism mitigates the risk of premature decoding for erratic tokens and thereby helps preserve overall generation quality.
    
% In this mechanism, $c_{fast}$ determines the allowable margin of the token's current confidence and its threshold.
%     $t_{start}$ means the warm-up time for the model.
%     The model may produce incorrect answers in the initial steps, but quickly starts the mainstream decoding after some steps. 
%     A long prompt often means a larger $t_{start}$. $c_{slow}$ determines the allowable margin of the confidence of the token and its threshold, and represents the aggression when decoding.

In this mechanism, $c_{\text{fast}}$ specifies the allowable margin between a token’s current confidence and its threshold. 
    $t_{\text{start}}$ denotes the warm-up period of the model. 
    In the initial steps, the model may generate incorrect answers, but it typically transitions to more reliable decoding behavior after sufficient iterations. 
    In practice, longer prompts often necessitate a larger $t_{\text{start}}$. 
    The parameter $c_{\text{slow}}$ similarly defines the allowable confidence and threshold margin while controlling the aggressiveness of the decoding process.

\subsection{Suspected $\text{Slow}$ Token Mechanism}

If a token is remasked with a confidence no more than $c_{slow}$ lower than its threshold, it is tagged with ``Suspected $\text{Slow}$'' label.
Once it accumulates the ``Suspected $\text{Slow}$'' label for $t_{max}$ consecutive steps, it is forced to decode.

In this mechanism, $t_{max}$ represents the patience that we have to wait for the token to exceed or drop below its threshold.
    A larger value of $t_{max}$ gives more freedom to the model while attenuating the effect of the threshold. 
    Similarly to $c_{fast}$, $c_{slow}$ specifies the minimum confidence required for a token to be considered as decodable. 
    In most cases, a value of approximately 0.05 is sufficient for those heavily-stalled tokens to be decoded properly.

The combination of confidence-threshold adjustment and Responsiveness optimization effectively integrates token confidence with its spatial-temporal dynamics, providing each token with an appropriate confidence threshold interval and, in turn, yielding a more reliable remasking strategy.
\section{\textbf{Experiment}}

% \begin{figure}[t]  % htbp是浮动位置参数
%     \centering  % 居中显示
%     \includegraphics[width=3.3in]{_fig/figures1.pdf}
%     \vspace{-2mm}
%     \caption{Radar charts of accuracy and speed using different remasking strategies}  % 图片标题
%     \vspace{-4mm}
%     \label{fig:7}  % 标签，用于交叉引用
% \end{figure}
% \begin{figure*}[!t]  % htbp是浮动位置参数
%     \centering  % 居中显示
%     \includegraphics[width=7in]{_fig/fig.pdf}
%     \vspace{-7mm}
%     \caption{Overview of Responsive Token Remasking for Diffusion Language }  % 图片标题
%     \vspace{-3mm}
%     \label{fig:7}  % 标签，用于交叉引用
% \end{figure*}
% ==============================================================================
% subsection
% ==============================================================================
\subsection{\textbf{Experiment Setup}}
We compare our method on four mainstream datasets: GSM8K~\cite{cobbe2021gsm8k}, MBPP~\cite{austin2021mbpp}, HumanEval~\cite{chen2021humaneval} and MATH500~\cite{hendrycksmath2021}. These datasets cover various programming scenarios and difficulty levels. We implemented and tested our method on the two mainstream open-source DLM models: LLaDA-Instruct-8B and Dream-7B.
Our experimental platform is based on 4*L40 GPUs. The generation length is set to 256 during the test. Our evaluation primarily considers accuracy, tokens per second second (TPS), and end-to-end latency. These metrics provide a comprehensive evaluation of the model's performance, capturing the inherent trade-offs between generation quality and inference speed.

For the parameters, we select $W_{t}=3$ and $W_{n}=3$; For MATH500, we select $W_{t}=5$ and $W_{n}=2$. In Responsiveness Optimization, we choose $t_{start} = 10$, $c_{fast}=c_{slow}=0.1$ and $t_{max}=3$ for all datasets. $p$ is set to 0.6 when the proportion of decoded tokens is below 20\% or above 80\%, and to 0.5 otherwise. More ablation studies about these parameters can be found in appendix.
% subsection
% ==============================================================================
\subsection{\textbf{Overall Performance}}
We compare our method with the baseline of LLaDA-Instruct-8B and Dream-7B, as well as the state-of-the-art inference acceleration method Fast-dLLM and another remasking strategy DUS.
Tab.~\ref{tab:1} presents the main results of our comparison on the GSM8K, MBPP, MATH500 and HumanEval datasets. 
% Fig.~\ref{fig:7} provides an intuitive visualization of our method on a series of datasets, showing that our method can adapt to various questions.
The experimental results consistently substantiate the superiority of our Responsive Token Remasking strategy over state-of-the-art methods across both LLaDA and Dream architectures. Notably, our approach achieves a dual breakthrough: it not only delivers substantial acceleration across all evaluated datasets but also enhances generative quality, effectively transcending the traditional speed-accuracy trade-off.

\textbf{\textit{Comparison in Speed}}:
When integrated with the LLaDA-8B model, STaRR consistently outperforms all baseline strategies. Notably, on the MBPP dataset, STaRR achieves a massive 8.90$\times$ acceleration over the Confidence Baseline. This far surpasses the 4.15$\times$ speedup of Fast-dLLM and the 2.70$\times$ of DUS, highlighting STaRR's exceptional capability in accelerating structured code generation tasks. Across GSM8K and MATH500, STaRR maintains this momentum with speedups of 3.07$\times$ and 3.74$\times$, respectively.

The robustness of our approach is further validated on the Dream-7B architecture. STaRR delivers substantial speedups of 3.41$\times$ on GSM8K and 3.65$\times$ on MATH500. Even on the high-complexity HumanEval benchmark, where DUS suffers from negative acceleration (0.61$\times$), STaRR still provides a solid 1.90$\times$ gain, confirming its scalability across diverse model architectures and task complexities. These results highlight that STaRR effectively mitigates the performance degradation often encountered by static strategies in logic-intensive tasks.
% The efficiency gains are even more pronounced in the throughput visualization. STaRR exhibits a substantial expansion in the TPS domain, outperforming both Fast-dLLM and DUS strategies by a significant margin.

\textbf{\textit{Comparison in Quality}}:
On the rigorous MATH500 benchmark, STaRR establishes a new performance peak with an accuracy of 35.1, surpassing the Confidence baseline (33.4) and significantly outperforming DUS (21.4). For Dream-7B on GSM8K, STaRR manages to reach a 75.0 accuracy, matching the high-fidelity Confidence baseline while being over three times faster.
In the MBPP task for Dream-7B, STaRR actually improves accuracy to 57.2 (vs. the baseline's 56.6), whereas other acceleration methods like Fast-dLLM (53.8) and DUS (46.4) experience noticeable performance degradation.
Unlike traditional methods that often sacrifice accuracy for speed, STaRR effectively achieves a Pareto improvement, positioning itself as a "plug-and-play" enhancement that safeguards the model's inherent reasoning capabilities while maximizing throughput.
% As shown in Fig.~\ref{fig:7} (a) and (c), the performance profile of STaRR closely aligns with the full-decoding baseline (Confidence), particularly on reasoning-intensive tasks such as GSM8K and MATH500. While traditional acceleration methods like DUS exhibit noticeable "craters" in specific domains (e.g., HumanEval), STaRR maintains a balanced and robust performance across varying sequence complexities. 

% ==============================================================================
% subsection
% ==============================================================================
\subsection{\textbf{Ablation Studies}}
% \begin{table}[!t]
%     \centering
%     \small
%     \caption{Influence of responsiveness Optimization \vspace{-2mm}}
%     \label{tab:2}
%     \begin{tabular}{c l c c c}
%         \toprule
%         ~ & \textbf{Method} & \textbf{Speedup $\uparrow$} & \textbf{Accuracy $\uparrow$}\\
%         \midrule
%         SOTA & Confidence & 1.0$\times$ & 79.2\\
%         \cmidrule{2-4}
%         \multirow{2}{*}{Ours} & \textbf{STaRR} w/ RO & 3.1$\times$ & 83.1\\
%         ~ & \textbf{STaRR} w/o RO & 1.98$\times$ & 79.2\\
%         \bottomrule
%     \end{tabular}
%     \vspace{-2mm}
% \end{table}
% \raggedbottom

\begin{table}[!t]
\centering
\scriptsize
\setlength{\tabcolsep}{4mm} % 消融实验列数少，可以适当加宽间距
\vskip -0.05 in
\caption{Influence of Responsiveness Optimization.}
\vskip -0.05 in
\begin{tabular}{c|c|cc}
\toprule
\toprule
\multirow{2}{*}{\textbf{Type}} & \multirow{2}{*}{\textbf{Method}} & \multicolumn{2}{c}{\textbf{Performance}} \\
\cmidrule{3-4}
~ & ~ & \textbf{Speedup $\uparrow$} & \textbf{Accuracy $\uparrow$} \\
\midrule
SOTA & Confidence & 1.00$\times$ & 79.2 \\
\midrule
\multirow{2}{*}{Ours} & STaRR w/o RO & 1.98$\times$ & 79.2 \\
~ & STaRR w/ RO & \textbf{3.10$\times$} & \textbf{83.1} \\
\bottomrule
\bottomrule
\end{tabular}
\label{tab:2}
\vskip -0.2 in
\end{table}

% \begin{table}[!b]
% \centering
% \scriptsize
% % \setlength{\tabcolsep}{1.5mm} # 调整表格间距
% \vskip -0.2 in
% \caption{Influence of Responsiveness Optimization}
% \vskip -0.05 in
% \begin{tabular}{c|c|c|c}
% \toprule
% \toprule

% \multirow{2}{*}{\textbf{Model}} & \multirow{2}{*}{\textbf{Environment}} & \multicolumn{2}{c|}{\textbf{VLA Inference}} & \multicolumn{2}{c}{\textbf{Database Retrieval}} \\
% \cmidrule{3-6}
% ~ & ~ & \textbf{SR} & \textbf{Speed} & \textbf{SR} & \textbf{Speed} \\
% \midrule
% \multirow{4}{*}{OpenVLA} & LIBERO-Goal & 77.0\% & 1.00$\times$ & 62.0\% & 4.17$\times$ \\
% ~ & LIBERO-Object  & 71.2\% & 1.00$\times$ & 68.0\% & 4.83$\times$ \\
% ~ & LIBERO-Spatial & 82.8\% & 1.00$\times$ & 53.0\% & 3.98$\times$ \\
% ~ & LIBERO-Long    & 54.4\% & 1.00$\times$ & 18.0\% & 3.74$\times$ \\
% \bottomrule
% \bottomrule
% \end{tabular}
% \label{tab:3-1}
% \end{table}
% \begin{table}[!t]
%     \centering
%     \small
%     \caption{Joint Utilization of dKV-cache and ICE \vspace{-2mm}}
%     \label{tab:3}
%     \begin{tabular}{c l c c c}
%         \toprule
%         ~ & \textbf{Method} & $\textbf{Speedup} \uparrow$  & $\textbf{Accuracy} \uparrow$\\
%         \midrule
%         SOTA & Confidence & 1.0$\times$ & 79.21\\
%         \cmidrule{2-4}
%         \multirow{4}{*}{Ours} & \textbf{STaRR} Only & 3.1$\times$ & 83.10\\
%         ~ & \textbf{STaRR} +dKV-Cache & 7.2$\times$ & 79.21\\
%         ~ & \textbf{STaRR} +ICE & 10.3$\times$ & 83.12\\
%         ~ & \textbf{STaRR} +dKV-Cache+ICE & 33.4$\times$ & 83.10\\
%         \bottomrule
%     \end{tabular}
%     \vspace{-2mm}
% \end{table}
% \raggedbottom

\begin{table}[!b]
\centering
\scriptsize
\setlength{\tabcolsep}{3.5mm} % 调整列间距，确保在单栏内美观
\vskip -0.05 in
\caption{Joint Utilization of dKV-cache and ICE.}
\vskip -0.05 in
\begin{tabular}{c|c|cc}
\toprule
\toprule
\multirow{2}{*}{\textbf{Type}} & \multirow{2}{*}{\textbf{Method}} & \multicolumn{2}{c}{\textbf{Performance}} \\
\cmidrule{3-4}
~ & ~ & \textbf{Speedup $\uparrow$} & \textbf{Accuracy $\uparrow$} \\
\midrule
SOTA & Confidence & 1.00$\times$ & 79.21 \\
\midrule
\multirow{4}{*}{Ours} & STaRR Only & 3.10$\times$ & 83.10 \\
~ & STaRR + dKV-Cache & 7.20$\times$ & 79.21 \\
~ & STaRR + ICE & 10.30$\times$ & 83.12 \\
~ & STaRR + dKV-Cache + ICE & \textbf{33.40$\times$} & \textbf{83.10} \\
\bottomrule
\bottomrule
\end{tabular}
\label{tab:3}
\vskip -0.2 in
\end{table}
\textbf{\textit{Influence of Responsiveness Optimization}}:
Responsiveness Optimization is an important stage of our method. We conduct experiments on the dataset GSM8K using LLaDA-Instruct-8B to analyze the role of Responsiveness Optimization played in the overall methodology. Table~\ref{tab:2} shows that applying Responsiveness Optimization can achieve a much higher speedup while guaranteeing the accuracy. The inclusion of RO propels the speedup from 1.98$\times$ to 3.1$\times$. Remarkably, it also yields a significant accuracy gain (from 79.2 to 83.1). In our experiments, we find that a larger number of suspected slow tokens were decoded by the method compared to suspected fast tokens being remasked in our experiment, which implies that Responsiveness optimization primarily serves as an accelerator.

\textbf{\textit{Integration with dKV-Cache and In-place Prompting}}:
In addition to optimizing remasking, there are many other DLM acceleration methods such as quantization\cite{zhang2025quant}, sparse-attention\cite{song2025sparse}, and speculative decoding~\cite{hong2025wide,li2025beyond,hong2025wide,wu2025specrouter,gao2025self}. 
STaRR can be adapted to the vast majority of existing methods.
dKV-cache ~\cite{ma2505dkv} stores and periodically updates the KV values of already generated tokens, whose KV values change little.
In-place prompting~\cite{jin2025thinking} (ICE) achieves more efficient, accurate generation by manually fixing certain tokens within the model's answer to force the model to think step by step.

We compared the changes in speedup and accuracy when our method is combined with dKV-Cache and In-place prompting on the dataset GSM8K. The results are shown in Table~\ref{tab:3}.
While individual mechanisms like dKV-cache and ICE provide incremental gains, their joint utilization with STaRR results in a synergistic surge in performance, reaching a peak speedup of 33.40$\times$. This exponential improvement underscores the orthogonality of our approach, demonstrating that STaRR can serve as a robust backbone for ultra-fast diffusion-based language generation without compromising output quality.

\textbf{\textit{Integration with Dual Cache}}:
Fast-dLLM adopts a block-based decoding scheme, which enables the use of cache. Dual Cache caches all Key-Value (KV) values external to the current block, making them available for its internal use. This serves as the primary acceleration mechanism for Fast-dLLM, yielding a $2\times$ to $3\times$ speedup in decoding without compromising generation quality.

Our framework offers the flexibility to be implemented in a block-based manner, where token dynamics are calculated exclusively using confidence levels within the local block. This localized computation, while slightly reducing the global context awareness, significantly facilitates memory efficiency and hardware-friendly execution. As shown in Table \ref{tab:5}, although the transition to a block-based strategy (Block STaRR) involves a slight trade-off in speedup ($2.88\times$ vs. $3.10\times$), the integration of the Dual Cache mechanism effectively compensates for this, boosting the cumulative speedup to $5.97\times$ with manageable accuracy variations.
\begin{table}[!t]
\centering
\scriptsize
\setlength{\tabcolsep}{3.5mm} % 调整列间距，确保在单栏内美观
\vskip -0.05 in
\caption{Block STaRR and Dual Cache}
\vskip -0.05 in
\begin{tabular}{c|c|cc}
\toprule
\toprule
\multirow{2}{*}{\textbf{Type}} & \multirow{2}{*}{\textbf{Method}} & \multicolumn{2}{c}{\textbf{Performance}} \\
\cmidrule{3-4}
~ & ~ & \textbf{Speedup $\uparrow$} & \textbf{Accuracy $\uparrow$} \\
\midrule
SOTA & Confidence & 1.00$\times$ & 79.21 \\
\midrule
\multirow{4}{*}{Ours} & STaRR & 3.10$\times$ & 83.10 \\
~ & Block STaRR & 2.88$\times$ & 79.21 \\
~ & Block STaRR + Dual Cache & 5.97$\times$ & \textbf{78.13} \\
\bottomrule
\bottomrule
\end{tabular}
\label{tab:5}
\vskip -0.2 in
\end{table}
\section{\textbf{Conclusion}}

% In this work, we reveal the intrinsic connection between token confidence and the model’s spacial-temporal dynamics, offering key insights into the decoding process of Diffusion Language Models. Building on this understanding, we introduce a responsive token-remasking mechanism, which is capable of ddressing the multifaceted challenges inherent in diffusion language model generation by adjusting confidence threshold for each token. Finally, empirical results demonstrate that our proposed approach strikes an optimal balance between inference speed and generation quality, outperforming baselines while maintaining seamless compatibility with existing frameworks.

We proposed \textbf{\textit{Spatial--Temporal Token-Dynamics-Aware Responsive Remasking}} (STaRR), a training-free framework that accelerates diffusion language model inference by exploiting the spatial--temporal dynamics of token confidence.
By quantifying temporal convergence and spatial correlations with lightweight metrics, STaRR enables a dynamic remasking strategy that better aligns with the model’s evolving semantic understanding.
Experiments show that STaRR achieves significant speedups while maintaining comparable accuracy, and remains complementary to existing inference acceleration techniques.
% ur study underscores the importance of token-level dynamics in diffusion-based generation, offering new insights for the design of efficient and principled decoding algorithms.

\newpage
\section{Impact Statements}
This paper presents work whose goal is to advance the field of diffusion language models. There are many potential societal consequences of our work, but none of which we feel must be specifically highlighted here.

% Note use of \abovespace and \belowspace to get reasonable spacing
% above and below tabular lines.

% \begin{table}[t]
%   \caption{Classification accuracies for naive Bayes and flexible
%     Bayes on various data sets.}
%   \label{sample-table}
%   \begin{center}
%     \begin{small}
%       \begin{sc}
%         \begin{tabular}{lcccr}
%           \toprule
%           Data set  & Naive         & Flexible      & Better?  \\
%           \midrule
%           Breast    & 95.9$\pm$ 0.2 & 96.7$\pm$ 0.2 & $\surd$  \\
%           Cleveland & 83.3$\pm$ 0.6 & 80.0$\pm$ 0.6 & $\times$ \\
%           Glass2    & 61.9$\pm$ 1.4 & 83.8$\pm$ 0.7 & $\surd$  \\
%           Credit    & 74.8$\pm$ 0.5 & 78.3$\pm$ 0.6 &          \\
%           Horse     & 73.3$\pm$ 0.9 & 69.7$\pm$ 1.0 & $\times$ \\
%           Meta      & 67.1$\pm$ 0.6 & 76.5$\pm$ 0.5 & $\surd$  \\
%           Pima      & 75.1$\pm$ 0.6 & 73.9$\pm$ 0.5 &          \\
%           Vehicle   & 44.9$\pm$ 0.6 & 61.5$\pm$ 0.4 & $\surd$  \\
%           \bottomrule
%         \end{tabular}
%       \end{sc}
%     \end{small}
%   \end{center}
%   \vskip -0.1in
% \end{table}

% In the unusual situation where you want a paper to appear in the
% references without citing it in the main text, use \nocite

\bibliography{ref.bib}
\bibliographystyle{icml2026}

%%%%%%%%%%%%%%%%%%%%%%%%%%%%%%%%%%%%%%%%%%%%%%%%%%%%%%%%%%%%%%%%%%%%%%%%%%%%%%%
%%%%%%%%%%%%%%%%%%%%%%%%%%%%%%%%%%%%%%%%%%%%%%%%%%%%%%%%%%%%%%%%%%%%%%%%%%%%%%%
% APPENDIX
%%%%%%%%%%%%%%%%%%%%%%%%%%%%%%%%%%%%%%%%%%%%%%%%%%%%%%%%%%%%%%%%%%%%%%%%%%%%%%%
%%%%%%%%%%%%%%%%%%%%%%%%%%%%%%%%%%%%%%%%%%%%%%%%%%%%%%%%%%%%%%%%%%%%%%%%%%%%%%%
\newpage
\appendix
\onecolumn
% \section{Case Study}
% \subsection{Evolution of Confidence and Threshold}

% \subsection{Confidence Distributions across Different Parts of Speech}
\section{Case Study}

\subsection{Evolution of Confidence and Threshold}

% 此处预留图片插入代码
% \begin{figure}[h]
%     \centering
%     \includegraphics[width=0.8\linewidth]{confidence_evolve.pdf}
%     \caption{Comparison of confidence trajectories and adaptive thresholds.}
%     \label{fig:confidence_evolution}
% \end{figure}

% \begin{figure*}[!t]  % htbp是浮动位置参数
%     \centering  % 居中显示
%     \includegraphics[width=7in]{_fig/Fig_7.pdf}
%     \vspace{-7mm}
%     \caption{Overview of Responsive Token Remasking for Diffusion Language }  % 图片标题
%     \vspace{-3mm}
%     \label{fig:6}  % 标签，用于交叉引用
% \end{figure*}

% \begin{figure*}[!t]  % htbp是浮动位置参数
%     \centering  % 居中显示
%     \includegraphics[width=7in]{_fig/Fig_8.pdf}
%     \vspace{-7mm}
%     \caption{Overview of Responsive Token Remasking for Diffusion Language }  % 图片标题
%     \vspace{-3mm}
%     \label{fig:6}  % 标签，用于交叉引用
% \end{figure*}

\label{app:evolution}

\begin{figure*}[ht]
    \centering
    % 第一张图：STaRR 方法
    \begin{subfigure}[b]{0.48\textwidth}
        \centering
        \includegraphics[width=\textwidth]{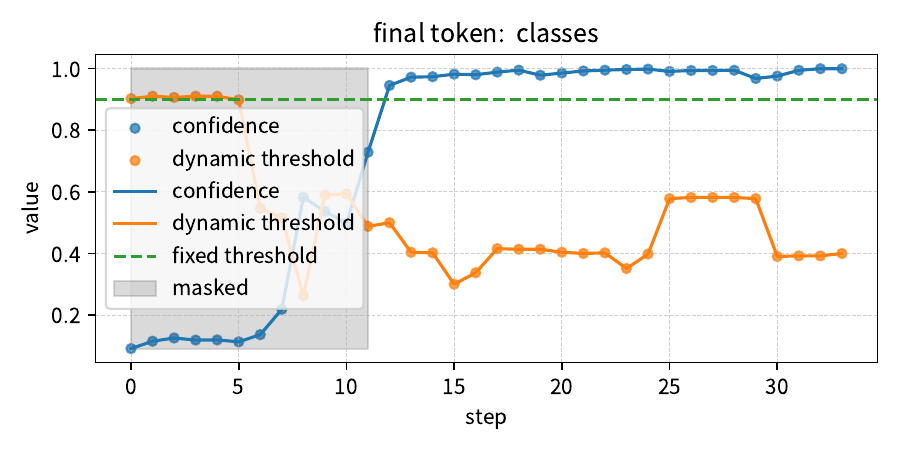}
        \caption{STaRR: Evolution of the token ``classes'' with responsive adaptive thresholding. Note how the dynamic threshold (orange) adapts to allow early unmasking at step 11.}
        \label{fig:starr_case}
    \end{subfigure}
    \hfill % 在两图之间插入弹性间距
    % 第二张图：Baseline 方法
    \begin{subfigure}[b]{0.48\textwidth}
        \centering
        \includegraphics[width=\textwidth]{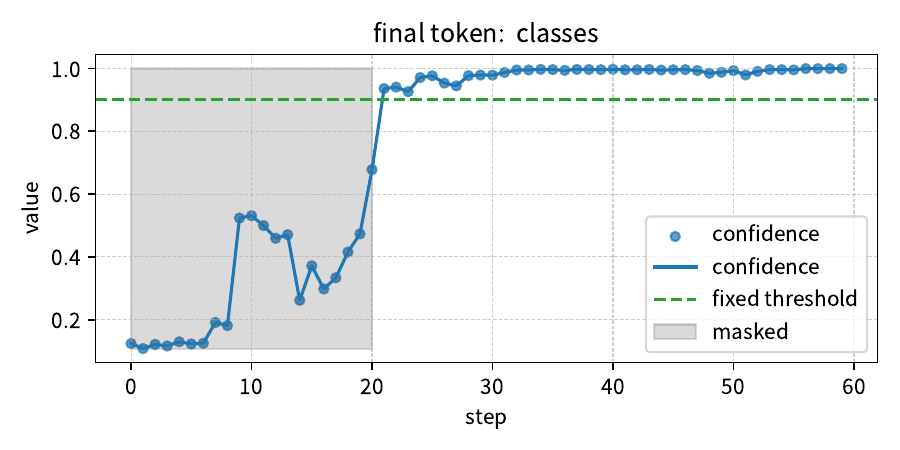}
        \caption{Fast-dLLM: Trajectory under a fixed 0.9 threshold. Despite the ``at-least-one'' decoding heuristic, the token remains masked until step 20, resulting in redundant iterations.}
        \label{fig:fix_case}
    \end{subfigure}
    
    \vspace{2mm} % 调节图与主标题的间距
    \caption{A comparative study of confidence trajectories and decoding thresholds. STaRR (left) identifies the point of sufficient certainty significantly earlier than the fixed-threshold baseline (right).}
    \label{fig:comparison_evolution}
    \vspace{-3mm} % 根据页面排版微调
\end{figure*}

To provide a granular view of the dynamic behavior of our \textit{Responsive Adaptive Thresholding},
we visualize the decoding trajectory of a representative token (``classes'') comparing STaRR (Fig.~\ref{fig:starr_case})
and the Fast-dLLM baseline (Fig.~\ref{fig:fix_case}).

% \paragraph{Dynamic Adaptation in STaRR.}
\noindent \textbf{Dynamic Adaptation in STaRR.} 
As illustrated in Fig.~\ref{fig:starr_case}, STaRR demonstrates a sophisticated balance between stability and efficiency.
During the initial five steps of the decoding process, the dynamic threshold $\tau_i^t$ is maintained at a high level (approximately 0.9).
This is attributed to the initialization of the $T_{base}$ at 0.95,
a strategic design choice that imposes a conservative constraint during the high-noise phase to
safeguard the semantic correctness of the early denoising. However, as the process progresses beyond this warm-up stage,
STaRR begins to respond to the token's internal convergence signals.
Around step 11, although the absolute confidence of ``classes'' oscillates near 0.7—significantly below the conventional 0.9 benchmark—
STaRR detects a stabilization in temporal variance and a favorable spatial consistency. 
Consequently, the threshold $\tau_i^t$ is adaptively lowered to approximately 0.5, 
allowing the token to be released from its masked state prematurely. 
% This early commitment enables the model to finalize the sequence by step 34, 
% demonstrating substantial speedup without compromising quality.

% \paragraph{Limitations of Static Confidence Decoding.}
\noindent \textbf{Limitations of Static Confidence Decoding.} 
In contrast, Fig.~\ref{fig:fix_case} highlights the inefficiency of the fixed-threshold approach used in Fast-dLLM.
Although Fast-dLLM implements a heuristic that mandates the decoding of at least one token with the highest confidence within each block per step,
this mechanism remains insufficient for tokens with slow-rising confidence.
The token ``classes'' fails to reach the static 0.9 threshold during the intermediate steps.
Since other tokens in the same block likely exhibited higher relative confidence,
``classes'' remained in a masked state, undergoing redundant iterations until it is unmasked by the ``at-least-one'' rule at step 20.
% This rigidity leads to a significantly protracted decoding process,
% with the sequence only reaching global convergence at step 60.

% \paragraph{Conclusion.} 
\noindent \textbf{Conclusion.} 
The comparison underscores that absolute confidence scores in DLMs are often overly conservative.
By leveraging spacial-temporal signals to adjust the threshold responsively—while maintaining a high-confidence anchor ($T_{base}=0.95$) 
in the initial steps—STaRR effectively identifies the point of sufficient certainty much earlier than methods tied to static thresholds or simple block-wise heuristics.

%do not remove
\begin{figure*}[!t]  % htbp是浮动位置参数
    \centering  % 居中显示
    \includegraphics[width=3.3in]{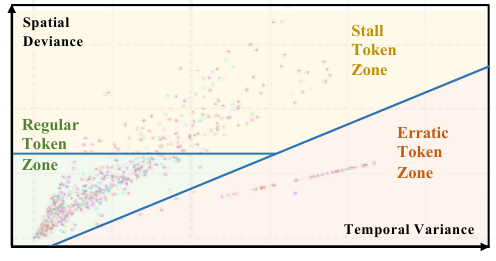}
    \vspace{-2mm}
    \caption{Dynamics Based Token Confidence Zone}  % 图片标题
    \vspace{-3mm}
    \label{fig:8}  % 标签，用于交叉引用
\end{figure*}
\subsection{\textbf{Dynamics Based Token Confidence Zone}}
The temporal and spatial features of token confidence are closely related. Fig.~\ref{fig:5} presents the distribution of each token’s temporal variance and spatial deviance across ten representative questions from the GSM8k dataset. Each point in this figure represents a single token.

Based on their temporal variance and spatial deviance, tokens can be partitioned into three functional zones. These zones are not strictly delineated; rather, they serve to describe the qualitative spatial-temporal characteristics of the tokens.

\noindent \textbf{Stall Zone.} 
Tokens with relatively low variance and high deviance falls into this zone. These tokens often exhibit a confidence gain at a specific iteration but remain below the threshold. Afterwards they become stable.
    So when their neighbors reach the confidence threshold they become different from them, thus get a high spacial deviance.

\noindent \textbf{Erratic Zone.} 
Tokens with relatively low deviance and high variance falls into this zone.
These tokens exhibit multiple confidence spikes. And they acquire abnormally high confidence scores during the early stages of decoding, which makes them outliers from neighbors.

\noindent \textbf{Regular Zone.} 
Tokens with low deviance and variance falls into this zone.
These tokens experience only a single confidence spike. It is only during this phase that a significant difference in confidence from their surrounding tokens is likely to emerge, making them more likely to be Consistent Tokens.

Traditional static remasking strategies generally fail to assign appropriate decoding opportunities to tokens in stall and erratic zones. This failure translates into erroneous decoding decisions for over 30 percent of the sequence—a flaw that proves catastrophic within the iterative diffusion process. Because suboptimal remasking selections in early stages significantly propagate and compound, they adversely affect subsequent decoding steps, ultimately preventing the model from achieving a Pareto-optimal balance between generation speed and linguistic quality.

% \subsection{Confidence Distributions across Different Parts of Speech}

% \section{Ablation Studies}
% \subsection{Datasets Information}

% \subsection{Respective Contributions of the Two Proposed Methods}

% \subsection{Sensitivity Analysis of Accuracy and Speed with respect to Window Size}
\section{More Ablation Studies}

\subsection{Datasets and Evaluation Metrics}
\label{app:datasets}

To provide a comprehensive overview of the experimental setup, 
we summarize the key statistics and evaluation protocols for the four benchmarks used in this study: 
GSM8K\cite{cobbe2021training}, HumanEval\cite{chen2021evaluating}, 
MBPP\cite{austin2021program}, and MATH500 (a representative subset of 500 problems sampled from the original MATH dataset\cite{hendrycks2021measuring}). 
These datasets are selected to evaluate the model's performance in high-precision reasoning and code generation tasks.

% \begin{table}[b]
%   \caption{Statistics and evaluation settings for the benchmarks used in our experiments.}
%   \label{tab:dataset-stats}
%   \begin{center}
%     \begin{small}
%       \begin{sc}
%         \begin{tabular}{lcccr}
%           \toprule
%           Dataset   & Task Type      & Test Size & Metric   & Setting \\
%           \midrule
%           GSM8K     & Math Reasoning & 1,319     & Accuracy & 5-Shot  \\
%           MATH500   & Advanced Math  & 500       & Accuracy & 4-Shot  \\
%           HumanEval & Code Synthesis & 164       & Pass@1   & 0-Shot  \\
%           MBPP      & Code Synthesis & 500       & Pass@1   & 3-Shot  \\
%           \bottomrule
%         \end{tabular}
%       \end{sc}
%     \end{small}
%   \end{center}
%   \vskip -0.1in
% \end{table}

\begin{table}[b]
\centering
\small % 1. 将字体从 \scriptsize 改为 \small (更大一级是 \normalsize)
\setlength{\tabcolsep}{3.5mm} % 2. 增加列间距，让表格更宽
\renewcommand{\arraystretch}{1.1} % 3. 稍微增加行高(默认是1.0)，让表格不那么拥挤
\vskip -0.05 in
\caption{Statistics and evaluation settings for the benchmarks used in our experiments.}
\vskip -0.05 in
\begin{tabular}{c|c|ccc}
\toprule
\toprule
\multirow{2}{*}{\textbf{Dataset}} & \multirow{2}{*}{\textbf{Task Type}} & \multicolumn{3}{c}{\textbf{Statistics}} \\
\cmidrule{3-5}
~ & ~ & \textbf{Test Size} & \textbf{Metric} & \textbf{Setting} \\
\midrule
GSM8K     & Math Reasoning & 1,319 & Accuracy & 5-Shot \\
MATH500   & Advanced Math  & 500   & Accuracy & 4-Shot \\
HumanEval & Code Synthesis & 164   & Pass@1   & 0-Shot \\
MBPP      & Code Synthesis & 500   & Pass@1   & 3-Shot \\
\bottomrule
\bottomrule
\end{tabular}
\label{tab:dataset-stats}
\vskip -0.2 in
\end{table}

% \paragraph{Mathematical Reasoning.} 
\noindent \textbf{Mathematical Reasoning.} 
For GSM8K and MATH500, we adopt the Chain-of-Thought (CoT) prompting method to elicit multi-step reasoning. The model's performance is measured by \textit{Exact Match} accuracy. We use a standardized rule-based parser to extract numerical answers from the generated trajectories. MATH500, as a representative subset of the MATH benchmark, involves competition-level problems requiring significantly higher logical consistency.

% \paragraph{Code Generation.} 
\noindent \textbf{Code Generation.} 
For HumanEval and MBPP, we evaluate the functional correctness of the generated Python scripts. We report the \textit{Pass@1} metric, which signifies the percentage of problems where the first generated solution passes all hidden unit tests. While HumanEval focuses on zero-shot hand-written problems, MBPP tests the model's ability to solve basic programming tasks with a few-shot setup.

\subsection{Respective Contributions of the Temporal and Spacial Thresholds}
\label{app:ablation_radar}

To further dissect the contributions of our proposed spacial-temporal components, we conduct a detailed ablation study on the GSM8K dataset. The results are visualized across two distinct diffusion language model backbones, LLaDA and DREAM, using radar charts that encompass three key dimensions: \textit{Accuracy}, \textit{Tokens Per Second} (TPS), and \textit{Remain Steps per sample} (representing the count of skipped decoding iterations).

\begin{figure*}[!t]
    \centering
    \begin{subfigure}[b]{0.48\textwidth}
        \centering
        \includegraphics[width=\textwidth]{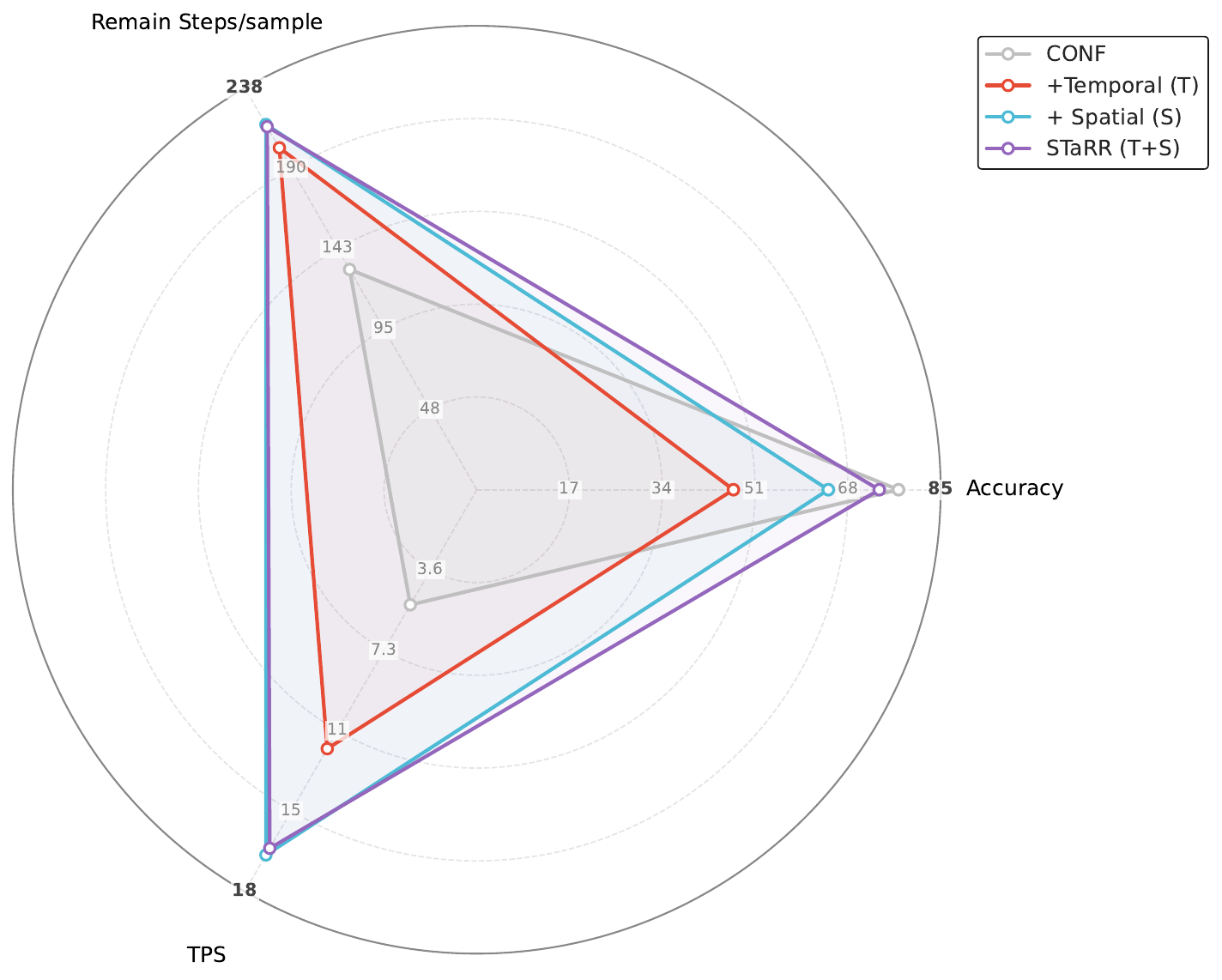}
        \caption{Ablation on LLaDA-8B.}
        \label{fig:ablation_llada}
    \end{subfigure}
    \hfill
    \begin{subfigure}[b]{0.48\textwidth}
        \centering
        \includegraphics[width=\textwidth]{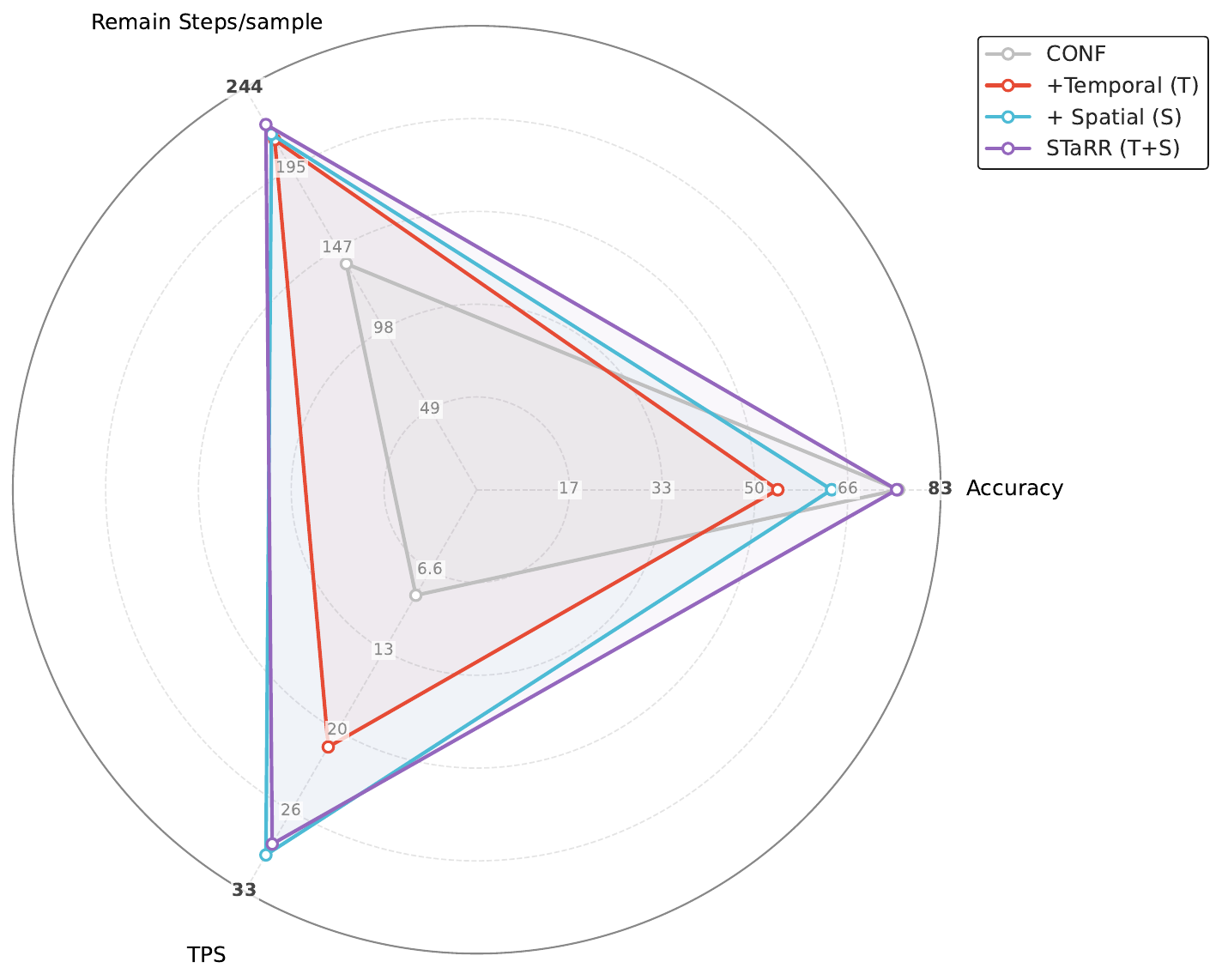}
        \caption{Ablation on Dream-7B.}
        \label{fig:ablation_dream}
    \end{subfigure}
    \vspace{2mm}
    \caption{Radar charts illustrating the performance on GSM8K. STaRR (T+S) consistently expands the performance envelope. Note that relying solely on temporal signals leads to a higher risk of decoding errors due to threshold decay for persistent masked tokens.}
    \label{fig:radar_comparison}
    \vskip -0.1 in
\end{figure*}

% \paragraph{Analysis of Component Synergy.} 
\noindent \textbf{Analysis of Component Synergy.} 
As illustrated in Fig.~\ref{fig:radar_comparison}, the \textbf{Temporal (T)} module (red) primarily drives efficiency gains, as evidenced by the expansion along the \textit{Remain Steps} axis. However, a critical limitation arises when relying exclusively on temporal variance: the threshold adjustment becomes strictly self-centric, depending only on a token's own historical states. In scenarios where a difficult token remains masked for an extended duration, the inherent decay of the confidence baseline ($c_{base}$) results in a progressively lower threshold. Without external contextual validation, this decay significantly heightens the probability of an erroneous token being prematurely decoded simply because its low-confidence state has stabilized over time.

% \paragraph{Mitigation via Spatial Consistency.}
\noindent \textbf{Mitigation via Spatial Consistency.} 
The \textbf{Spatial (S)} module (cyan) acts as a crucial corrective mechanism to this temporal vulnerability. By integrating local semantic consistency, the spatial deviance metric ensures that a token is not released solely based on its temporal stability but also its compatibility with the surrounding context. On the GSM8K benchmark, this synergy is particularly vital for multi-step reasoning. In both LLaDA (Fig.~\ref{fig:ablation_llada}) and DREAM (Fig.~\ref{fig:ablation_dream}), the \textbf{STaRR (T+S)} variant (purple) forms the largest convex hull, demonstrating that spatial signals provide a necessary ``semantic anchor'' that prevents the efficiency-oriented temporal module from compromising logical accuracy.

\subsection{Sensitivity Analysis of Accuracy and Speed with respect to Window Size}
\label{app:sensitivity}

To gain a deeper understanding of how the spatial-temporal window sizes influence the adaptive remasking process, we conduct a sensitivity analysis on the hyperparameters $W_t$ (temporal window size) and $W_n$ (spatial window size). For this study, we select a representative subset of 100 samples from the GSM8K dataset to perform a grid search over varying window scales. While this sampled subset allows for a granular observation of the trade-off between accuracy and the \textit{Average Number of Function Evaluations} (NFE), it should be noted that the limited sample size introduces a degree of stochasticity in the results, leading to minor non-monotonic fluctuations in the observed curves.

\begin{figure*}[ht]
    \centering
    \begin{subfigure}[b]{0.48\textwidth}
        \centering
        \includegraphics[width=\textwidth]{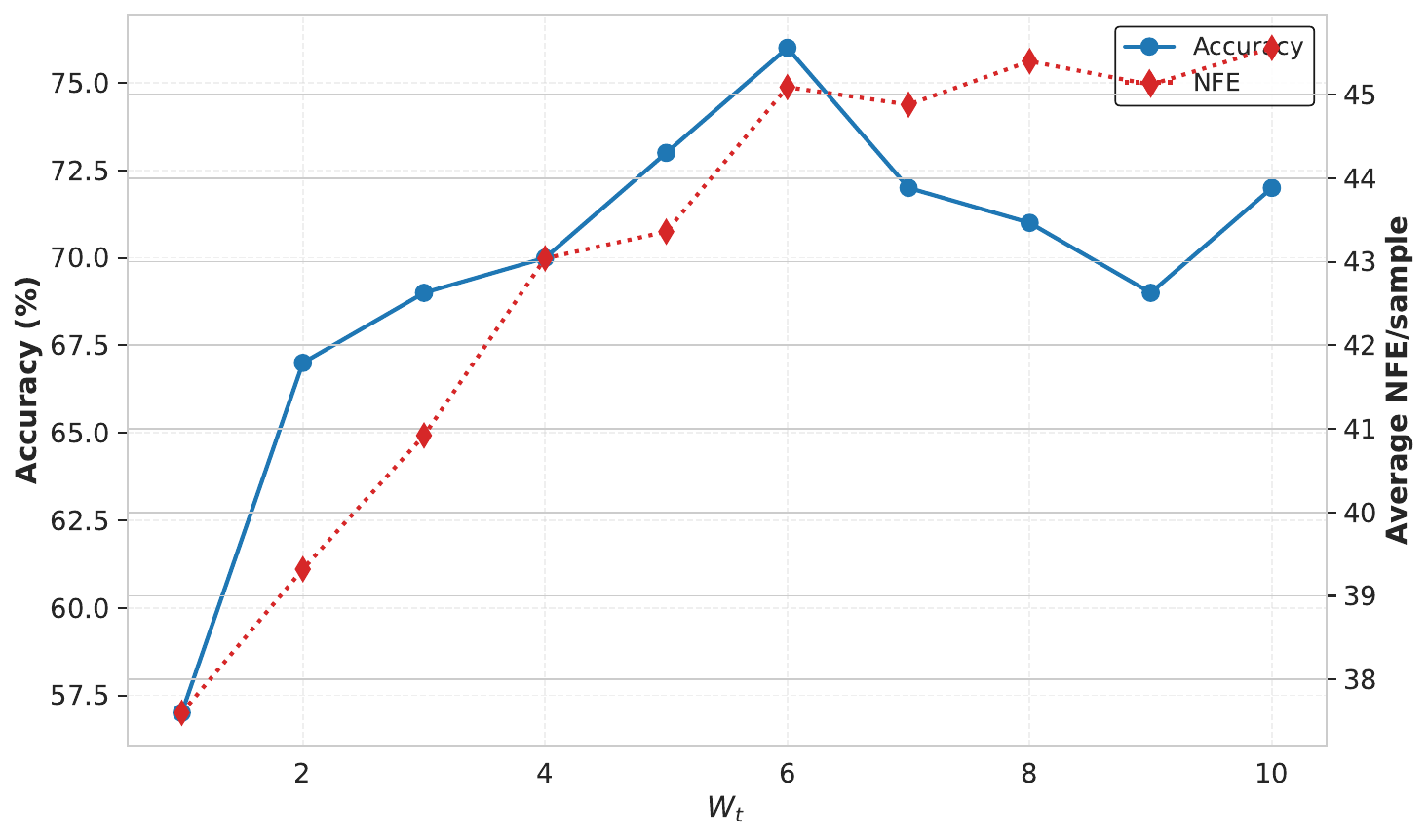}
        \caption{Impact of temporal window size $W_t$.}
        \label{fig:sensitivity_wt}
    \end{subfigure}
    \hfill
    \begin{subfigure}[b]{0.48\textwidth}
        \centering
        \includegraphics[width=\textwidth]{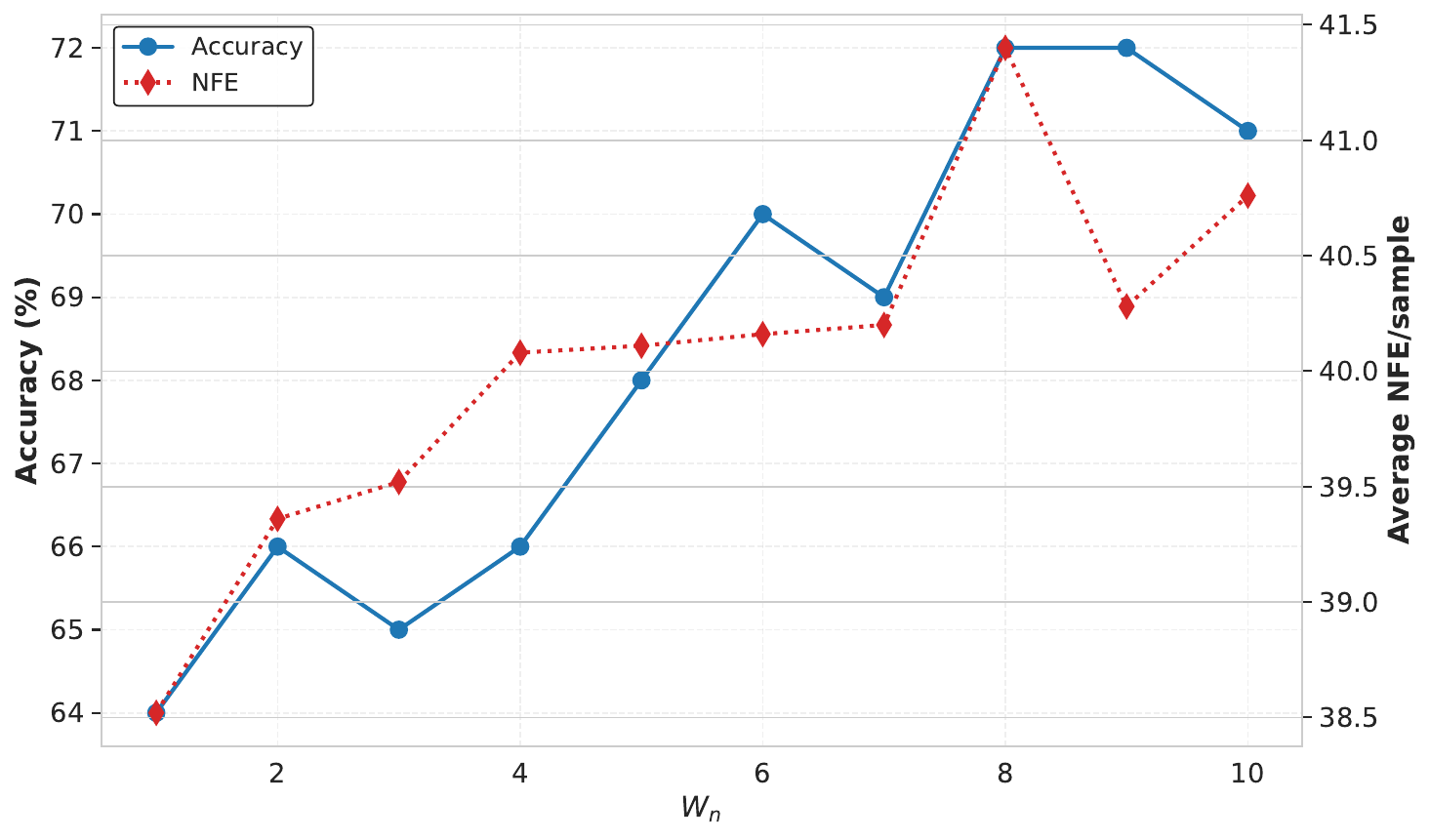}
        \caption{Impact of spatial window size $W_n$.}
        \label{fig:sensitivity_wn}
    \end{subfigure}
    \vspace{2mm}
    \caption{Sensitivity analysis of spacial-temporal window sizes on a 100-sample subset of GSM8K. (Left) Accuracy peaks at $W_t=6$, with NFE generally increasing as the temporal history expands. (Right) Expanding the spatial context generally benefits reasoning accuracy, though the curves exhibit minor noise due to the stochastic nature of the small-scale evaluation.}
    \label{fig:sensitivity_analysis}
    \vskip -0.1in
\end{figure*}

% \paragraph{Analysis of Temporal Window $W_t$.} 
\noindent \textbf{Analysis of Temporal Window $W_t$.} 
As illustrated in Fig.~\ref{fig:sensitivity_wt}, $W_t$ is instrumental in stabilizing the temporal variance signal $var_t^i$. At $W_t=1$, the threshold is highly susceptible to transient noise, causing premature unmasking and low accuracy. As $W_t$ increases to 6, the accuracy improves sharply to its peak, confirming that a moderate history is essential for reliable convergence estimation. Beyond $W_t=6$, we observe a gradual increase in NFE alongside slight accuracy fluctuations. This pattern suggests that while a larger window provides a more robust variance estimate, it also introduces a ``decision lag'' that potentially delays the unmasking of stable tokens. The minor dip observed at $W_t=9$ is likely a localized artifact of the stochasticity inherent in the 100-sample test set.

% \paragraph{Analysis of Spatial Window $W_n$.}
\noindent \textbf{Analysis of Spatial Window $W_n$.} 
Fig.~\ref{fig:sensitivity_wn} depicts the influence of the spatial neighborhood $W_n$ on the \textit{Spatial Deviance} metric. The accuracy demonstrates an overall upward trend as $W_n$ scales from 1 to 8, underscoring the importance of contextual consistency in identifying semantic errors. Although the NFE and accuracy curves exhibit some jaggedness—attributable to the randomness of the small-scale evaluation—the optimal performance is consistently observed when $W_n$ is between 8 and 10. The relative stability of NFE across different $W_n$ values indicates that spatial checking provides a high-efficiency mechanism for refining remasking decisions without incurring significant computational overhead.

%%%%%%%%%%%%%%%%%%%%%%%%%%%%%%%%%%%%%%%%%%%%%%%%%%%%%%%%%%%%%%%%%%%%%%%%%%%%%%%
%%%%%%%%%%%%%%%%%%%%%%%%%%%%%%%%%%%%%%%%%%%%%%%%%%%%%%%%%%%%%%%%%%%%%%%%%%%%%%%

\end{document}